\documentclass[11pt,a4paper]{article}
\usepackage[hyperref]{naaclhlt2019}
\usepackage{times}
\usepackage{latexsym}

\usepackage{url}

\usepackage{xspace}
\usepackage{xargs} 
\usepackage[table]{colortbl}

\usepackage{booktabs}

\usepackage{array}
\usepackage{graphicx}
\usepackage{subfig}

\usepackage{amsmath}
\definecolor{OliveGreen}{cmyk}{0.64,0,0.95,0.40}
\usepackage{comment}

\usepackage{bbding}
\usepackage{pifont}
\usepackage{wasysym}
\usepackage{amssymb}
\usepackage{amsthm}
\usepackage{mathtools}

\usepackage{commath}
\usepackage{pdflscape}
\usepackage{tabularx,array}

\usepackage{enumerate}
\usepackage{enumitem}

\usepackage{algpseudocode}

\usepackage{times}
\usepackage{fancyhdr}
\usepackage{multirow}

\usepackage[
  separate-uncertainty = true,
  multi-part-units = repeat,
  detect-weight = true 
]{siunitx}
\usepackage{booktabs}
\usepackage{etoolbox,siunitx}
\robustify\bfseries
\usepackage{array,multirow,graphicx}

\usepackage{array}
\newcolumntype{P}[1]{>{\centering\arraybackslash}p{#1}}

\usepackage{xcolor}

\usepackage{setspace}

\usepackage[ruled,vlined]{algorithm2e}

\SetCommentSty{mycommfont}
\SetKwInput{KwInput}{Input}                
\SetKwInput{KwReturn}{return}

\usepackage[colorinlistoftodos,prependcaption,textsize=tiny]{todonotes}

\usepackage{marginnote}

\newcommandx{\siva}[2][1=]{\todo[linecolor=red,backgroundcolor=red!10,bordercolor=red,#1]{SR: #2}\xspace}
\newcommand{\ignore}[1]{}

\usepackage{ragged2e,array,booktabs}
\usepackage{bbding}
\usepackage{pifont}
\usepackage{wasysym}
\usepackage{amssymb}
\usepackage{balance}
\usepackage{pbox}

\aclfinalcopy 

\title{On the Importance of Karaka Framework in Multi-modal Grounding}
\author{
  Sai Kiran Gorthi \\
  IIIT Hyderabad \\
  {\tt } \\
  Radhika Mamidi \\
  IIIT Hyderabad \\
  {\tt radhika.mamidi@iiit.ac.in} \\
  }

\date{}

\begin{document}
\maketitle
\begin{abstract}
Computational Paninian Grammar (CPG) model helps in decoding a natural language expression as a series of modiﬁer-modiﬁed relations and therefore facilitates in identifying dependency relations closer to language/context semantics compared to the usual stanford dependency relations. However, the importance of this CPG dependency scheme has not been studied in the context of multi-modal vision and language applications. At IIIT-H, we plan to perform a novel study to explore the potential advantages and disadvantages of CPG framework in a vision-language navigation task setting, a popular and challenging multi-modal grounding task. 

\end{abstract}

\section{Introduction}
The Computational Paninian Grammar (CPG) framework, inspired by inflectionally rich language Sanskrit~\cite{bharati1996paninian,bharati1993parsing}, provides the level of syntactico-semantic intepretation of a sentence and has been successfully applied to several tasks~\cite{gupta2012novel,akula2013novel,akula2015novel,palakurthi2015classification}. Karaka based relations are generated by this framework are known to facilitate relations between verbs and other related syntactic constituents such as nouns in a language expression/sentence. More concretely, CPG treats a language expression as a series of modifier-modified relations~\cite{begum2008dependency}. This CPG parsing framework has been shown to be very effective for morphologically richer languages with relatively free word order such as Indian languages. In addition, there is a prior work extending CPG framework to English resulting in an elegant computational grammar~\cite{vaidya2009karaka}. However, the benefits and limitations of this framework has not been recently studied in the context of multi-modal grounding problems~\cite{antol2015vqa,akula2022question,akula2019natural}.

At IIIT-H, we critically examine the importance of CPG based karaka relations in solving the Vision-and-Language Navigation (VLN) task, a fundamental and challenging multi-modal problem~\cite{fried2018speaker,anderson2018vision}. VLN is an emerging research area aimed at building embodied agent that can communicate with humans through natural language expressions in real 3D environments. Specifically, the VLN agent navigates and interacts with the surrounding objects in the scene to complete the task according to the given input language instructions. Recent works showed that state-of-the-art models fail at comprehending the underlying semantics of language instructions and rather look at salient words such as nouns. We believe that this is one of the primary reason for observing a very low navigation performance on these datasets such as ALFRED~\cite{shridhar2020alfred}. We plan to conduct a study to evaluate whether if the modifier-modified based syntactico-semantic karaka relations in CPG could help in grounding the deeper semantics of language expressions in VLN.

To achieve this, we pick the ET model~\cite{pashevich2021episodic}, a top-performing VLN model on ALFRED benchmark and plug in karaka annotations during pre-training the language tower. We compare the performance of this approach with the parts-of-speech based pre-training, syntax tree based pre-training and dependency based pre-training (see Section 2). We will draw several interesting insights with this by conducting extensive ablations and analysis on the generalization and interpretation capabilities~\cite{akula2019explainable} of ET model with all these different language tower pre-training strategies. 

\begin{figure*}[t]
\centering
  \includegraphics[width=0.95\linewidth]{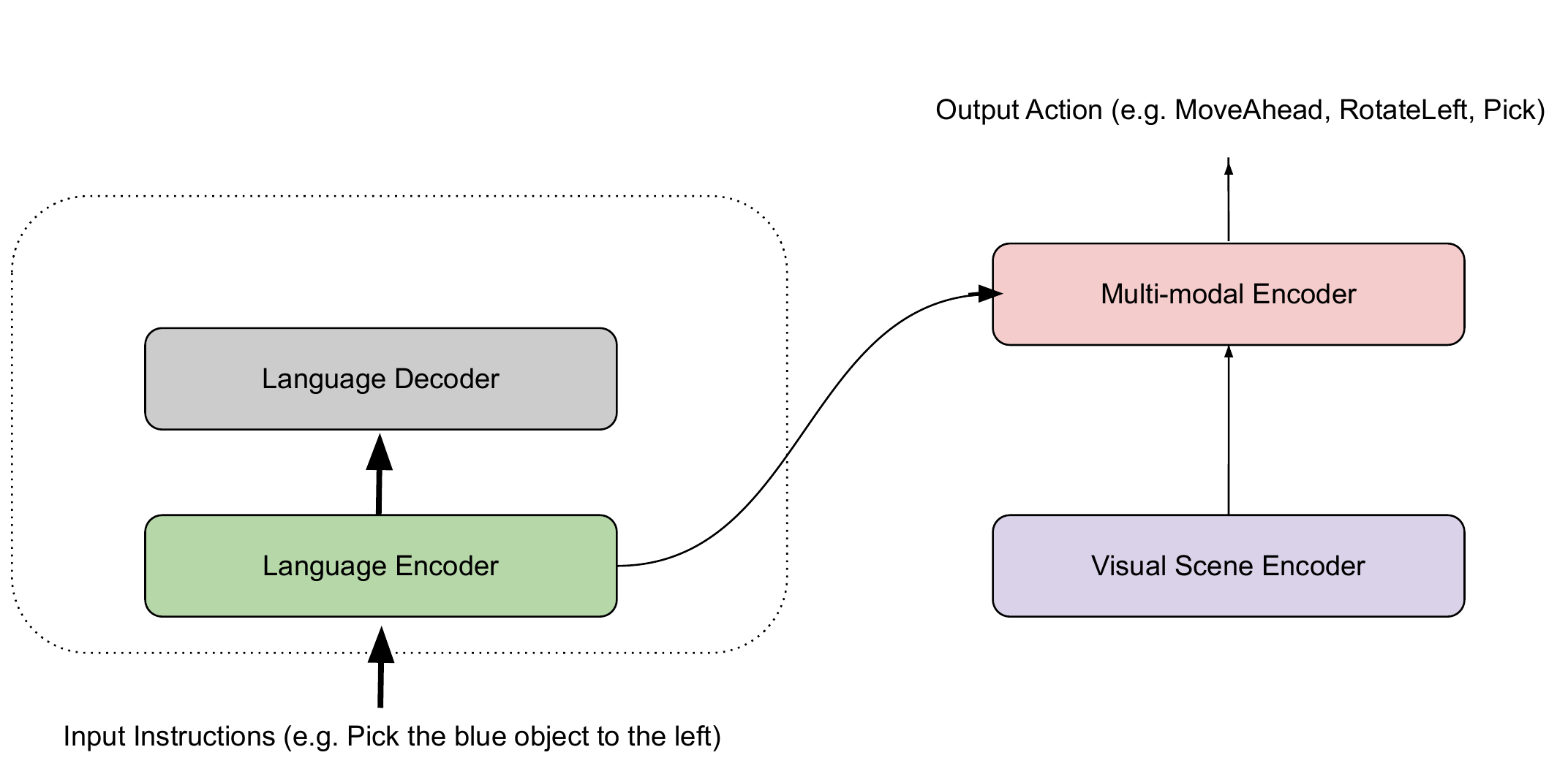}
  \caption{ET model for ALFRED: Separate encoders for encoding representations of language and visual information; and an additional multi-modal encoder to attend over both language and visual representations}~\label{fig:intro1}
\end{figure*}

\begin{figure*}[t]
\centering
  \includegraphics[width=0.95\linewidth]{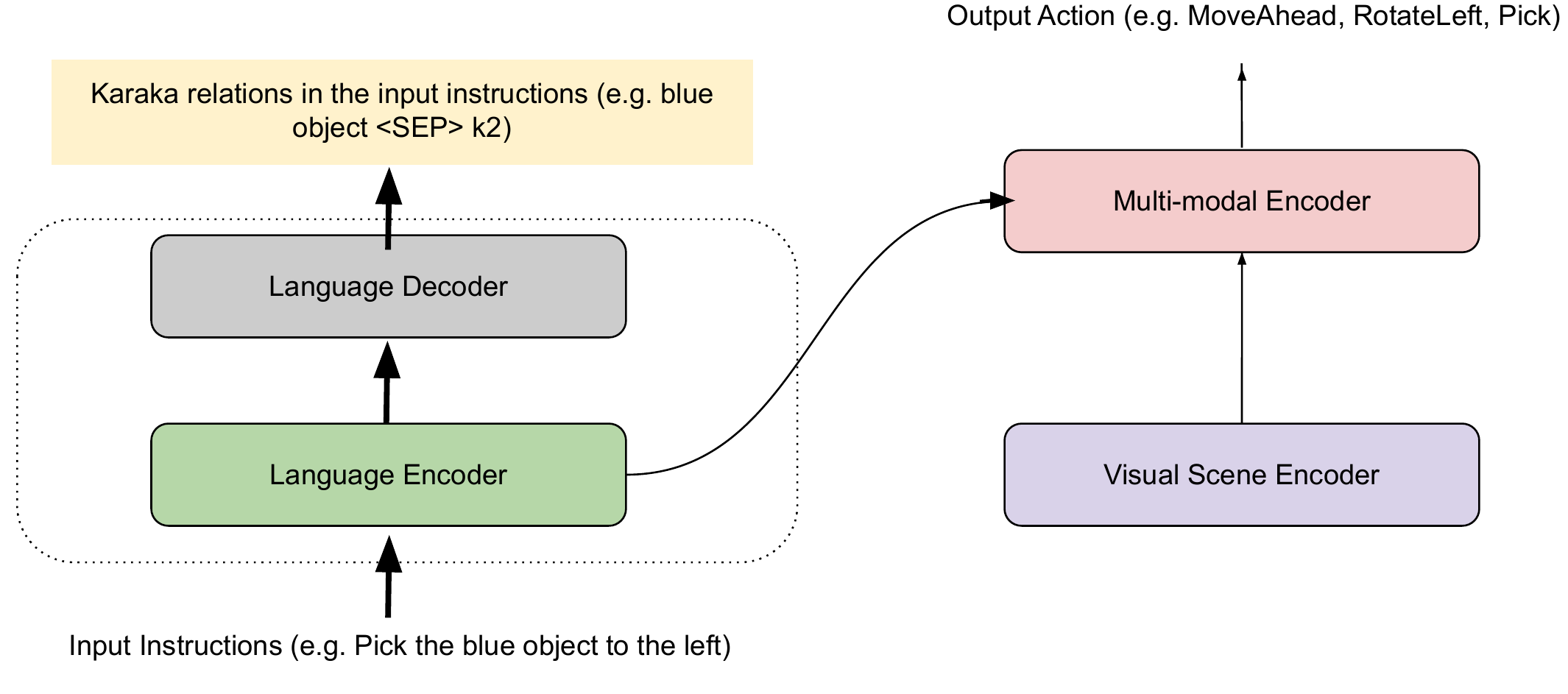}
  \caption{Our Approach: Enforcing language encoder to explicitly predict karaka relations to facilitate in better grounding of spatial relations in the input language instructions.}~\label{fig:intro2}
\end{figure*}

\section{Importance of Karaka Relations in Multi-modal grounding}
An embodied AI agent needs to strictly comprehend the language instructions to be able to successfully solve the interaction and navigation tasks. As discussed in prior works~\cite{zhang2020diagnosing,zhu2021diagnosing}, it is sometimes possible for these agents to ignore language instructions and just rely on visual images in navigating the environment. This is problematic as this fails to test the language understanding abilities of these agents. ALFRED~\cite{shridhar2020alfred} benchmark introduces a complex object interaction and navigation environment where it is relatively more difficult to solve the task successfully without comprehending language instructions. This is because, ALFRED consists of longer sequence of instructions with diverse set of target objects. In this work, We focus on the ALFRED environment and its defined tasks.

Transformer based architectures~\cite{vaswani2017attention}, have been shown to be successful in solving a wide range of multi-modal grounding problems including VLN~\cite{akula2021crossvqa,akula2021robust,akula2020cocox,akula2022effective}. In particular, they efficiently map long horizon tasks to the target set of actions. For our analysis, we choose the ET model~\cite{pashevich2021episodic}, a top-performing Transformer based VLN model on ALFRED leaderboard. ET model relies on attention-based multi-layer transformer encoders and observes the full history of visual observations and previous actions (in predicting next next action). More specifically, ET has separate encoders for encoding representations of language and visual information; and an additional multi-modal encoder to attend over both language and visual representations~\cite{akula2019visual,gupta2016desire,akula2019x,akula2018analyzing}. The language encoder takes the language instructions as input and outputs a sequence of contextualized language embeddings. The visual encoder projects a sequence of visual observations into its embedding. The multimodal encoder is a multi-layer transformer encoder, outputs a new embedding by taking the concatenated language and visual embeddings from the modality-specific encoders. Figure~\ref{fig:intro1} demonstrates the high-level modules in ET. 

Out goal is to improve the performance of ET model, which suffers greatly on navigation sub-tasks~\cite{pashevich2021episodic,zhu2021diagnosing}, by explicitly enforcing the spatial relations from the input instructions in the language encoder pre-training. To do this, we plug in the CPG karaka annotations during pre-training the language encoder in ET. We compare the performance of this approach with the other syntax based enforcing methods such as parts-of-speech based pre-training, syntax tree based pre-training and dependency based pre-training. This allow us to observe several interesting insights by conducting extensive ablations and analysis on the generalization capabilities of ET model with all these different language tower pre-training strategies. We demonstrate our approach in Figure~\ref{fig:intro2}.

\section{Conclusion}
In this paper, we proposed our plan to critically examine the importance of CPG dependency scheme in the context of grounding semantics of language in multi-modal environment. Through our novel study, we explored the potential advantages and disadvantages of CPG framework in solving ALFRED tasks, a popular and challenging VLN benchmark.


\end{document}